# Individual Tree Detection and Crown Delineation with 3D Information from Multi-view Satellite Images

Changlin Xiao, Rongjun Qin, Xiao Xie, and Xu Huang

## Abstract

*Individual tree detection and crown delineation (ITDD) are critical in forest inventory management and remote sensing based forest surveys are largely carried out through satellite images. However, most of these surveys only use 2D spectral information which normally has not enough clues for ITDD. To fully explore the satellite images, we propose a ITDD method using the orthophoto and digital surface model (DSM) derived from the multi-view satellite data. Our algorithm utilizes the top-hat morphological operation to efficiently extract the local maxima from DSM as treetops, and then feed them to a modified superpixel segmentation that combines both 2D and 3D information for tree crown delineation. In subsequent steps, our method incorporates the biological characteristics of the crowns through plant allometric equation to falsify potential outliers. Experiments against manually marked tree plots on three representative regions have demonstrated promising results – the best overall detection accuracy can be 89%.*

## Introduction

Forest is one of the most important land surfaces and plays an important role in the global ecosystem. Timely and accurate measurements of the forest parameters at the individual tree level such as tree count, tree height, and crown size are essential for quantitative analysis of forest structure, ecological modeling, biomass estimation, and evaluation of deforestations (Mohan *et al.*, 2017; Weng *et al.*, 2015). Over the past several decades, remote sensing techniques have greatly improved the capability of extracting forest metrics with high spatial resolution imagery (Gonçalves *et al.*, 2017; Sousa *et al.*, 2015). However, the majority of these methods use spectral or texture information and are limited to the radiometric quality which makes them vulnerable to erroneous detections, either over-/under- segmenting tree crowns at the individual plot level.

The addition of 3D information to these forest metric estimations can greatly enhance the measurement accuracy. Recently, much attention has been given to lidar data, which provides an accurate 3D representation of the surface objects. A number of algorithms have been proposed to analyze the forest structure at individual tree level with this data (Ferraz *et al.*, 2016; Kathuria *et al.*, 2016; Liu *et al.*, 2015). Many of the methods that are based on either lidar or photogrammetric 3D points use the normalized digital surface model (nDSM) or the canopy height model (CHM, for the forest applications), which can naturally highlight the treetops and directly offer the tree heights (Liu *et al.*, 2015; Lu *et al.*, 2014). The CHM can be generated by subtracting the digital terrain model (DTM) from the digital surface model (DSM). Similar to the 2D image-based methods, the CHM-based methods use procedures such as image smoothing, local maximum localization, and template matching to detect the individual trees and their boundaries (Koch *et al.*, 2006; Popescu *et al.*, 2002). In Strîmbu and Strîmbu (2015), they used graph theory to model the forest topological structure and correct two potentially over-identified treetops. Also, multi-scale segmentations have been proposed to dynamically select the best set of apices and generate the final segmentation (Véga *et al.*, 2014). For tree crown delineation, image segmentation methods such as valley following, region growing, and watershed segmentation can be directly used on CHM (Ferraz *et al.*, 2016; Kathuria *et al.*, 2016; Strîmbu and Strîmbu, 2015). Among these methods, the watershed segmentation is the most popular as it can naturally and efficiently model the treetops and crowns, for example, Liu *et al.* (2015) proposed the Fishing Net Dragging (FND) method which uses the watershed segmentation with the Gaussian filtering to find the boundaries of trees. Even though these methods have demonstrated great successes in their applications, they are limited by the cost of lidar surveying making them impractical for repetitive acquisition of at large scales (Wulder *et al.*, 2013). Also, lidar systems offering compensatory spectrometers are not widely available to provide additional spectrum data for more advanced analysis, such as LAI (leaf area index), NDVI (normalized difference vegetation index) for vegetation classification. A demand for such data normally requires an additional light for multi-/hyper-spectral data acquisition while which subsequently brings in registration issues.

Changlin Xiao is with the Future Cities Laboratory, Singapore-ETH Centre, ETH Zurich, 1 Create Way, CREATE Tower, #06-01, 138602, Singapore; and the Department of Civil, Environmental and Geodetic Engineering, The Ohio State University, 218B Bolz Hall, 2036 Neil Avenue, Columbus, OH 43210.

Rongjun Qin is with the Department of Civil, Environmental and Geodetic Engineering, The Ohio State University, 218B Bolz Hall, 2036 Neil Avenue, Columbus, OH 43210; and the Department of Electrical and Computer Engineering, The Ohio State University, 205 Dreese Labs, 2015 Neil Avenue, Columbus, OH, 43210 (qin.324@osu.edu).

Xiao Xie is with the Research Center for Industrial Ecology & Sustainability, Institute of Applied Ecology, Chinese Academy of Sciences. No.72, Wenhua Road, Shenhe District, Shenyang City 110016, China; and the Key Laboratory for Environment Computation & Sustainability of Liaoning Province. No.72, Wenhua Road, Shenhe District, Shenyang City 110016, China.

Xu Huang is with the Department of Civil, Environmental and Geodetic Engineering, The Ohio State University, 218B Bolz Hall, 2036 Neil Avenue, Columbus, OH 43210.

Considering the merits of both 2D spectral and 3D structure data, for the first time, we propose to use multi-view high-resolution satellite imagery to perform forest parameters retrieval at the individual tree level. With the growing number of high-resolution satellite sensors, the chances of a spot being viewed multiple times with multiple angles are greatly increased. These multi-view images can facilitate many remote sensing tasks, for example, Liu and Abd-Elrahman (2018) used multi-view images in a deep convolutional neural network for the wetlands classification. Also, the development of the advanced image matching algorithms makes it possible to produce comparably dense 3D measurements as lidar, while with much lower cost, higher flexibility in acquiring information in a large geographical region. With the highly accurate 3D digital surface models (DSM) and true orthophotos generated from these multi-view satellite images, we expect it will significantly enhance the performance of individual tree detection and crown delineation (ITDD).

The terrain data is critical for many 3D points based ITDD methods. However, for DSM generated from satellite images, terrain data under the forest canopy might not be captured. Moreover, considering that the accuracy of image-based DSM is normally lower than those from lidar, it can be particularly challenging to directly use the point cloud based methods on this DSM. Hence, to fully explore the multi-view satellite imagery based data, we propose a novel algorithm that utilizes both 2D spectral and 3D structural information with the orthophoto and DSM. Based on the assumption that tree crowns are normally well rounded in shape with a single maximum as the treetops, we propose to use morphological top-hat by reconstruction (THR) to detect treetops. Compared to other local maximum detectors, for example, window-based local maximum filters (Pouliot and King, 2005; Wulder *et al.*, 2000), the THR detector is less sensitive to the window or filter size. For the crown delineation, we adopt a modified superpixel segmentation framework to generate compact segments that leverage the boundary of crowns based on the DSM and multispectral information, thus are able to account for crown delineation in both sparsely and densely forested area. Compared to the previous methods, for instance, valley following, region growing, and watershed segmentation (Gougeon and Leckie, 2006; Ke and Quackenbush, 2011), the modified superpixel segmentation is similar to region growing, but with an extra spatial constraint which ensures more compact shapes for trees. The tree crowns are complex in their 3D structure, including overlapping canopies, adjacent crowns reflecting similar spectrums but different in height, and smooth crowns. Hence, the combination of both 2D spectral and 3D structural information would greatly help to identify the individual trees.

To the authors' best knowledge, this work contributes to the community as the first to demonstrate and offer the use of multi-view high-resolution satellite imagery as an alternative data source for the forest parameter retrieval at the individual tree level. In addition, the contributions include the development of a novel top-hat and superpixel based detection framework that is able to (1) accommodate multi-modal data for segmenting objects under complex scenarios; (2) utilize biometric characteristics of trees through the allometric equation to constrain size and shape of tree segments; and (3) achieve high accuracy in areas with different canopy densities. In particular, the proposed method is able to account for densely forested regions, even without a high precision DTM.

## Study Area and the Data Processing
The study area is located in Don Torcuato, a small city on the west side of Buenos Aires, Argentina. In this area, we choose three experimental sites with tree plots at different levels of density as illuminated in Figure 1 and Figure 2. Site A, covering an area of *0.30 × 0.30* km², which we treat as the sparsely forested area mainly consists of sparsely distributed trees, wild shrubs, and grasslands. In the densely forested site B, covering *0.25 × 0.34* km², different types of trees at different heights intersect with each other and only a small part of it being ground surface. Site C is part of a small town, covering an area of *0.30 × 0.30* km², the surface objects of which are complicated: trees at the courtyard or around buildings with different crown sizes and heights; shrubs on the street sides mixed with trees at different heights.

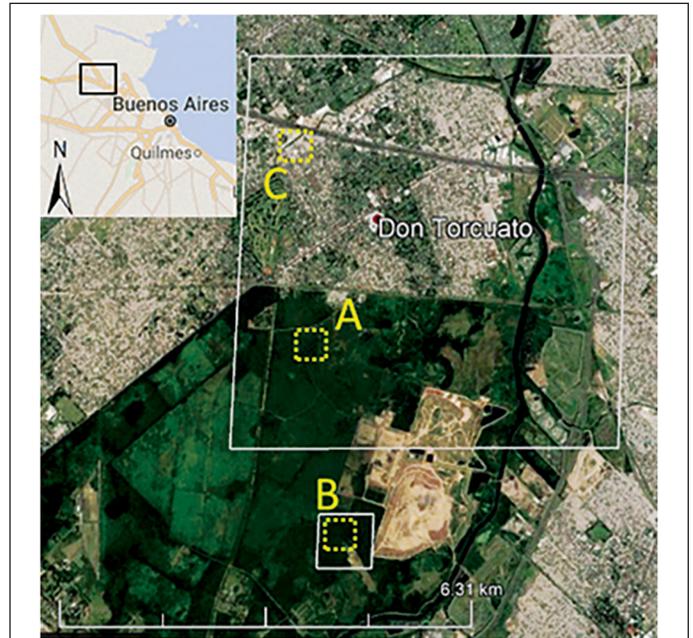

Figure 1. The study area near Buenos Aires, Argentina. The two large solid rectangles mark the areas where we have generated the DSM and orthophoto and A, B, C mark the experimental sites.

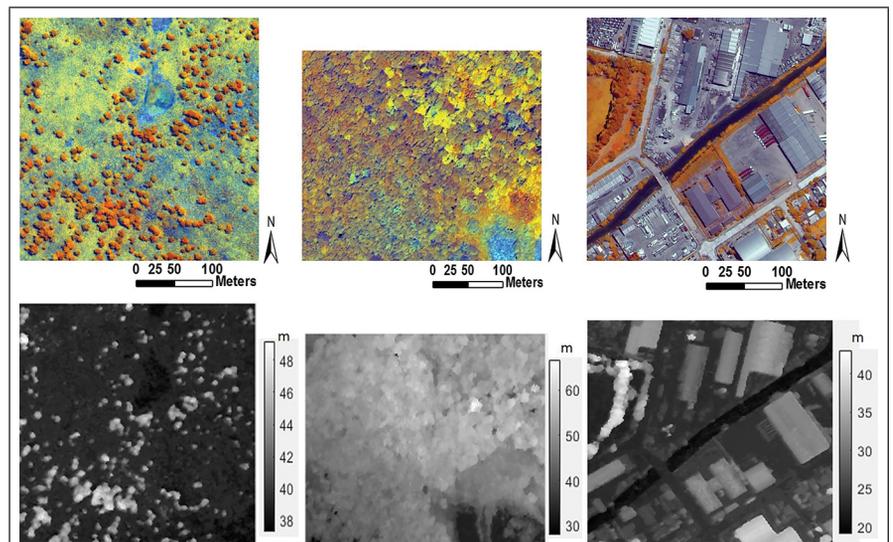

Figure 2. The three experimental sites in the study area. From left to right are the orthophoto (upper) and the DSM (lower) of Site A, B, and C with different surfaces.

The satellite images in this work are from the multi-view benchmark dataset provided by John's Hopkins University Applied Physics Lab (JHUAPL) (Bosch *et al.*, 2016; Bosch *et al.*, 2017). The data contains 8 bands Worldview2/3 images with the ground resolution around *0.3* meters. To derive an accurate DSM, we selected five pairs of the on-track stereo images captured in December 2015, with the maximal off-nadir angle between *7-19* degrees and the average intersection angle between *15-21* degrees. We applied a fully automated pipeline (Qin, 2017) that consists of (1) pansharpening, (2) automatic feature matching, (3) pair-wise bundle adjustment, (4) dense matching, and (5) a bilateral-filter based depth-fusion, to generate a high-quality DSM and subsequently true orthophoto. Comparing to the ground truth lidar data, the root-mean-square errors (RMSE) of the DSM are varying between *2.5-4* meters which is absolute accuracy at checking points which do not represent the relative accuracy of the object reconstruction. More details about the method and the accuracy evaluation can be found in Qin (2017) and Figure 2 shows the cropped orthophoto and DSM of the three experimental sites.

## Methodology

The proposed method includes several steps summarized in Figure 3: After the generation of an orthophoto and DSM, the vegetated area and terrain area are extracted to facilitate the treetop detection which is based on the local maximum of DSM through top-hat by reconstruction (THR) operation (Qin and Fang, 2014). To further improve the detection quality, we use above ground height check and non-maximum suppression with the allometric equation to eliminate the short and redundant detections. From the treetops, a modified superpixel segmentation that combines the 2D spectral and the 3D structural information is proposed to effectively delineate the tree crowns. Finally, a postprocessing for the crown refinement is used to further improve the detection accuracy.

### Vegetation and Terrain Detection

To identify the vegetation areas, we use the Normalized Difference Vegetation Index (NDVI) as the index and take the areas where their NDVI > μ to be the vegetation area. μ is empirically set as *0.3* leveraged based on our experiments. Digital terrain model (DTM) is a useful source for the individual tree detection, and there have been several methods proposed to extract the DTM from 3D points (Gevaert *et al.*, 2018; Hu *et al.*, 2014). Such as in Hu *et al.* (2014), they proposed an adaptive surface filter (ASF) that the threshold can vary according to the terrain smoothness to efficiently classify the airborne laser scanning data. DTM is used to offer the height information as normalized DSM (nDSM) or CHM (for forest application). However, in some densely forested areas, it may not be feasible to extract the DTM from DSM produced by images. Fortunately, the proposed method is not heavily depended on the tree height information. The treetop detection and crown delineation are mainly decided by the relative height. The absolute tree height is an extra cue to refine detections which will be discussed with more details in the experiments.

Our method does not explicitly generate the nDSM or CHM. Instead, we focus on the height gradients and estimate the above ground tree height with an effective terrain detection method. By converting the pixels in DSM map as grid point cloud, we apply the cloth simulation filter (CSF) (Zhang *et al.*, 2016) that based on the height and surrounding information to classify the points into two categories: terrain and off-terrain points. The above-ground heights of trees which are subsequently used for crown size estimation can be estimated by subtracting the terrain height around it, and more details can be found in the next section. In the experiments, we used the open source software CloudCompare® as the segmentation tool, and an example can be found in Figure 4.

### Treetop Detection

*Detection of Local Maximum Points*
Similar to lidar-based treetop detection, we naturally assume that the local maximum in the DSM is the treetop. However, since the filter-based method requires a careful tuning of window size, we adopt the grey-level morphological top-hat by reconstruction operator (THR) to find the local maximum, as it is an effective method of detection blob-like shapes and less sensitive to window size (Qin and Fang, 2014). In the detection, a disk-shaped structuring element (SE) is used to perform the grey-level morphology erosion on the DSM to generate a marker image $\varepsilon$(DSM, $e$), where the erosion operation only keeps the minimum value of all the pixels in the structuring element. The morphological reconstruction mask $B_{\varepsilon(DSM, e)}$ is then generated through an iterative procedure in which the dilation operation that keeps the maximum value is utilized on the marker image. Finally, by subtracting $B_{\varepsilon(DSM, e)}$ from the DSM, the peaks on the DSM can be extracted as blob-shaped peak regions with respect to the local maximum.

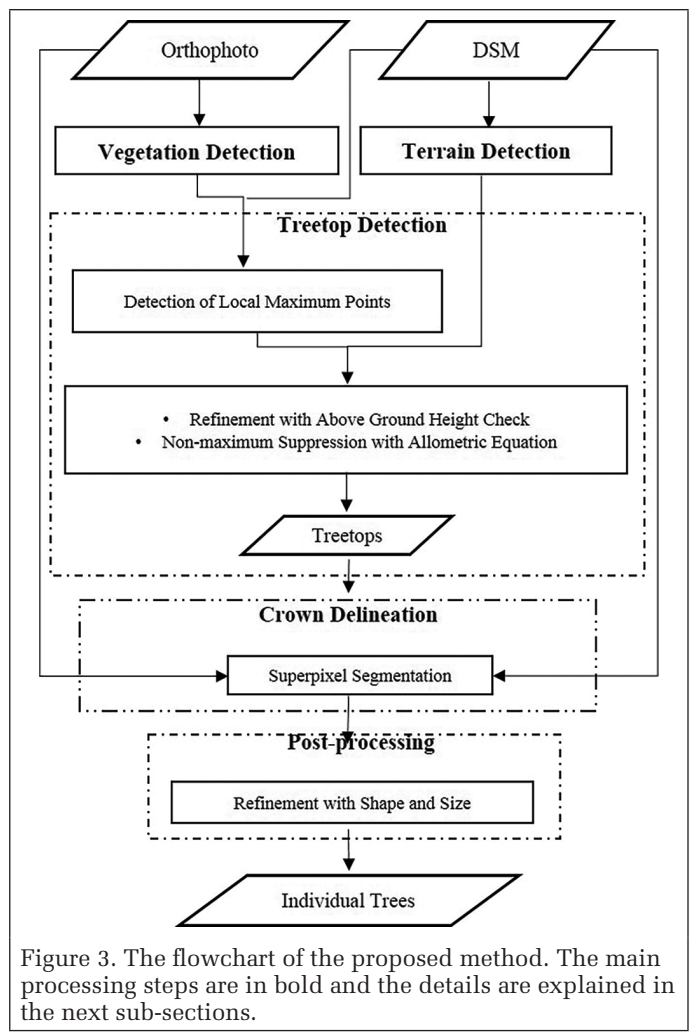

Figure 3. The flowchart of the proposed method. The main processing steps are in bold and the details are explained in the next sub-sections.

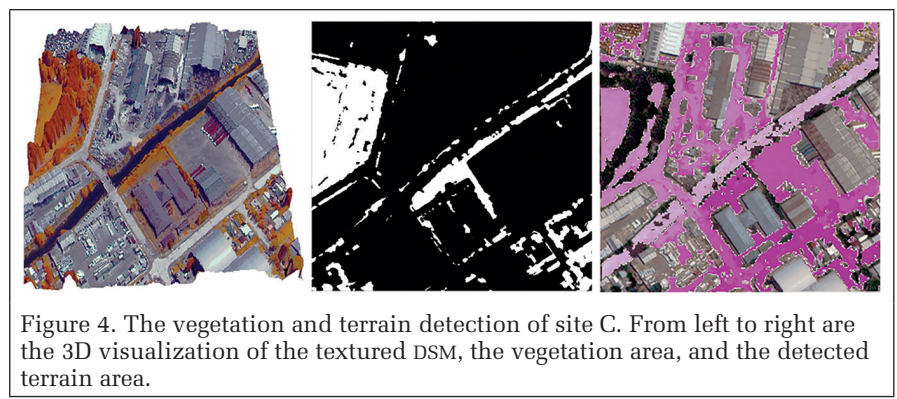

Figure 4. The vegetation and terrain detection of site C. From left to right are the 3D visualization of the textured DSM, the vegetation area, and the detected terrain area.

To locate the local maximum at the pixel level, we first use the morphological opening operation to remove the weakly connected parts thus to separate a big region into several small ones. Then, for each region, we keep the highest point as the final treetop candidate. Figure 5 shows an example of the local maximum detection with two SEs (*4* and *8* pixels). The detected local maximum regions with different SE sizes are shown as white and red dots (with a shift) in the rightmost image of Figure 5. As we can observe they are almost identical indicating the proposed THR operation is less sensitive to the size of SE.

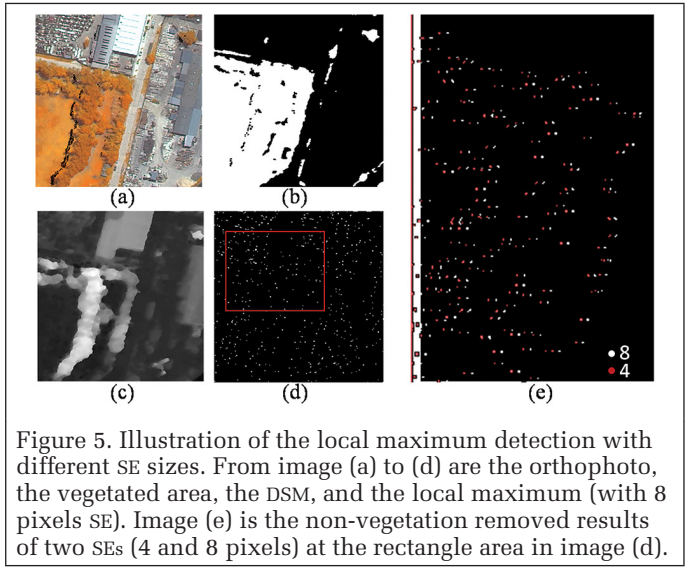

Figure 5. Illustration of the local maximum detection with different SE sizes. From image (a) to (d) are the orthophoto, the vegetated area, the DSM, and the local maximum (with 8 pixels SE). Image (e) is the non-vegetation removed results of two SEs (4 and 8 pixels) at the rectangle area in image (d).

*Refinement with Above Ground Height Check and Non-Maximum Suppression*
To eliminate lower local maximum, we calculate per-point above ground height by subtracting its DSM height with nearby terrain height:

$$H(s) = DSM(s) - \sum_{p_i \subseteq A_i, p_i \subseteq T} DSM(p_i)/N , \qquad (1)$$

where $H(s)$ is the above ground height of point $s$; DSM($s$) and DSM($p$) are the height values at point $s$ and $p$. $N$ is the total number of the terrain points in the predefined search window $A_i$ centered at $s$, and $T$ is the extracted terrain area as previously described.

To remove the redundant maximum points in one tree, a common practice is to use non-maximum filters to locate the true maximum within a window. The filter is expected to achieve the best performance when the window size is close to the crown size. However, the crown size of different types of trees may vary significantly. Given that the allometric equations describe the biological relationships between the tree height and its crown size (Garrity *et al.*, 2012), it can be of a great value to use such cue to determine an optimal window size accountable for crown variations. While in general larger window is able to encapsulate small crowns, we consider to a type with a large crown/height ration to obtain a good estimate as for removing non-maximum. Hence, we adopt the allometric equation for the deciduous tree (Desktop, 2011) to estimate the crown size (or filter window size) $\chi$ for each treetop as:

$$\chi(s_i) = 3.09632 + 0.00895 * h_t(s_i)^2 , \qquad (2)$$

where $h_t$ represents the above ground height of the treetop $s_i$. Figure 6 gives an example of the refinement of the local maximum (red dots), many (yellow circle marked) of which are not treetops to the final refined treetops (stars with a blue dot in the center). Also, in the same figure, the rectangles mark the non-maximum suppression window for each potential treetop and we can observe that they are associated with a sizeable window which is able to account for crown size variation.

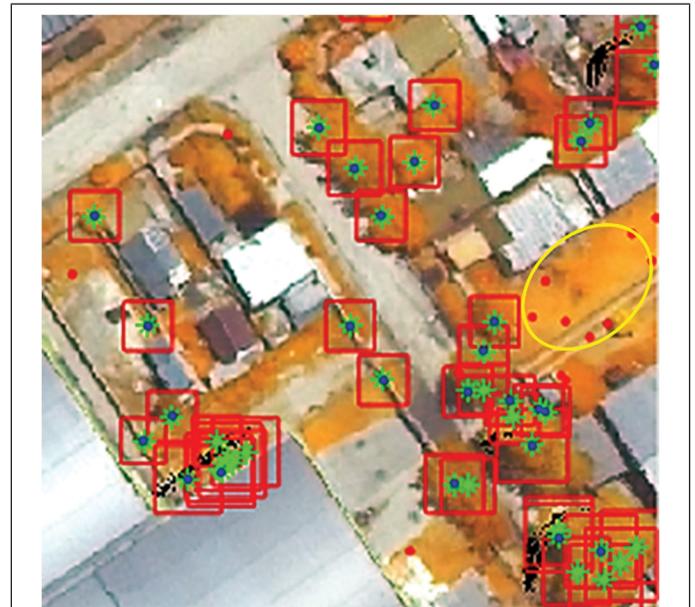

Figure 6. Treetop refinement. The initial detections are filtered by the above ground height check and non-maximum suppression with adaptive windows.

**Crown Delineation**
The crown boundaries are usually indistinct in canopy density area and the texture information is normally not sufficient for the crown delineation. To utilize the 3D structure information of the DSM, we propose a modified superpixel segmentation utilizing multi-modal data to detect crown regions. The superpixel algorithm can provide a compact and homogenous representation towards all segments which naturally matches the shape of trees. In contrast to the original superpixel (Felzenszwalb and Huttenlocher, 2004), the modified version considers both 2D and 3D information in its kernel distance function:

$$D(s_i, t_j) = W_h * D_h(s_i, t_j) + W_v * D_v(s_i, t_j) + W_c * D_c(s_i, t_j) \qquad (3)$$

where $D(s_i, t_j)$ measures the spatial and spectral difference between treetop $s_i$ and the test pixel $t_j$. $D_h$, $D_v$, and $D_c$ are both the Euclidean distance used to measure the horizontal, vertical distance and the spectral differences (*8* bands) and normalized based on the statistical maximum and minimum range of tree crown size (estimated by allometric equation), tree height, and the vegetation spectrum. $W_h, W_v, W_c$ are the associated weights for different component which we empirically set as *0.8, 1, 0.5*, emphasizing spatial distance especially the vertical one. The lower $D(s_i, t_j)$ means higher similarity between the treetop and the test pixel, but if the smallest difference between a pixel and all the treetops is still larger than a certain threshold value $\theta$, the pixel will be abandoned as non-tree area. The $\theta$ gives an extra constraint on the assignment of pixels that may not belong to trees. Smaller $\theta$ regulates the delineated crown to be more compact and closer to the treetop, but may lose part of the true crown. On the contrary, larger $\theta$ would reduce the restriction and let the delineation cover a larger area which may contain non-tree part. Hence, in the experiment, the $\theta$ is carefully selected as the maximum difference value the Equation 3 could have, for example, the difference between the treetop and the leaf at the farthest crown. The idea of the revised superpxiel is similar to the supervoxels (Papon *et al.*, 2013) used for full 3D data like those from terrestrial lidar. However, unlike the full 3D points, the DSM only represents 2.5 D information that only the surface points have height information. If using supervoxels, the facades (which were assumed to be cut-off planes) will be considered for segmentation which is unwanted. Besides, we have fixed the segmentation seeds at the treetops and the goal is to create a boundary delineation in the planar map. Hence, it is more appropriate to use this modified superpixel than the supervoxels.

**Postprocessing Using the Crown Shape and Size Constraints**
In this study, we set two criteria to further verify the correctness of the crown's shape and size. The first criterion is that the treetop should be near the center of the crown. In the experiment, the one-third of the largest diameter of the segment is set as the maximum tolerance range. The falsely detected treetops that are far from the segment center are often the local maximum at the edge of the crowns and they are usually caused by the limited precision of the DSM. The other criterion is the coherence of the crown size. Normally, a tree and its neighbors belong to the same species and should share common biological features (e.g., the height-crown ratio). Therefore, we use the average crown size of the neighborhood as the reference crown size to remove abnormal crowns with the three-sigma rule.

## Experiments and Discussion
**Accuracy Assessment Measure**
To quantitatively validate the individual tree detection and crown delineation accuracy, we use true positives (TP), false positives (FP) and false negatives (FN) to compute the detection accuracy (DA) and recall (r), commission error ($e_{com}$) and the omission error ($e_{om}$):

$$r = DA = \frac{n_{TP}}{N}, \quad e_{com} = \frac{n_{FP}}{n_{TP} + n_{FP}}, \quad e_{om} = \frac{n_{FN}}{n_{TP} + n_{FN}}, \quad (4)$$

where $n_{TP}$, $n_{FN}$ and $n_{FP}$ are the number of trees in TP, FN and FP category. $N$ is the total number of the reference trees. Also, the precision ($P$) and F-score($F$) are derived as:

$$p = \frac{n_{TP}}{n_{TP} + n_{FP}}, \quad F = \frac{2rp}{r + p}. \quad (5)$$

These are normally effective in pixel-wise comparison or detection of trees in the sparsely vegetated area. However, it might be problematic when we are validating the algorithm in the densely vegetated area: one predicted tree may have several reference trees nearby. Thus, it is hard to pair the predictions and references. To enforce a one-to-one correspondence, we wish only match the pair that has the largest overlapping area to each other. Therefore, following the measurement employed by Pascal visual object classes (VOC) challenge (Everingham *et al.*, 2010), we calculate the overlap ratio (OR) between all reference and predicted tree crowns to estimate how well they matched:

$$OR = \frac{2 * A_o}{A_r + A_p}, \quad (6)$$

where $A_o$, $A_r$, and $A_p$ are the size of overlapped, reference and prediction crown, respectively. However, if the corresponding trees have small overlap ratio than $\gamma$, we will discard this pair. We adopt $\gamma=0.3$ as the threshold as it used in Yin and Wang (2016). For cases that: (1) one with no corresponding reference, it will be counted as a false positive; (2) one reference tree does not correspond to a predicted tree, it will be counted as a false negative. Finally, for the crown delineation accuracy, we estimate the average overlap ratio of the matched pairs:

$$CA = \frac{\sum OR(TP_i)}{n_{TP}}, \quad (7)$$

where $CA$ is the crown accuracy, $OR(TP_i)$ is the overlap ratio of correctly matched reference and prediction crowns.

## Results
In this work, we are limited to collect field samples and we generated the reference data by labeling the individual trees and their crowns with visual inspection as some previous studies did in their works (Zhen *et al.*, 2014). The three experimental sites variably include densely forested area, sparsely forested area, and urban area, respectively. For each site, we calculated the detection accuracy (DA), commission error($e_{com}$), omission error ($e_{om}$) and crown accuracy $CA$, as well as the precision $P$ and F-scores.

To demonstrate the advantage of the top-hat local maximum detector that is less sensitive to filter size, we implemented several local maximum filter based treetop detectors described in Wulder *et al.* (2000). These include fixed-window filters with sizes of *3,7,11,15,19* pixels corresponding to *1-6* meters and the filter with a variable size calculated by the slope breaks (SB) (Wulder et al., 2000). In the experiment, we test these filters on the DSM in our comparative study while keeping all the other steps as the same and the final results can be found from Table 1 to Table 3.

As shown in Table 1 to Table 3, the proposed top-hat detector has the majority of highest performances across the three test sites. We believe this is due to the top-hat's characteristics of robustness and filter size insensitivity. For the fixed-window filters, the window with *7* pixels has better performance in the sparsely vegetated area, while the window with *11* pixels has better performance in the other two test sites. This shows that the performance of a fixed-window is dependent on the compatibility of the window size and the scenario. However, a suitable filter size cannot be predicted, and it is not possible to find a single filter suitable for all scenarios. And in general, the larger windows miss more small trees resulting in higher omission errors. Compared to the fixed-window detector, the proposed top-hat detector produces reliable treetop in all scenarios independent of the SE size. On the other hand, the variable-window filter based on the slope break obtains the worst results. This could be due to that the slope break distance is sensitive to its nearby environment and cannot accurately reflect the crown size.

To understand the improvement of including 3D geometric information, we performed the treetop detection and superpixel segmentation without DSM. Following the idea in Wulder *et al.* (2000), we use the red channel which has the best performance with the same detection processes excluding the 3D information. In Equation 3, the vertical distance is removed and the other parameters are tuned to obtain the optimal results which can be found in Table 4. Since the spectral-only based method soley considers the reflection variations of light off of trees, it is difficult to distinguish trees from other objects. Even with the help of NDVI which can remove the non-vegetation area, the grassland and bushes can still be treat as part of the trees. As we can observe the results in the sparsely forested area and the urban area in Table 4, both of them have significantly high commision error ($e_{com}$) due to the grasslands being identified as forests. Also, without DSM, all the performances have significally decreased which demonstras the importance of including the 3D information.

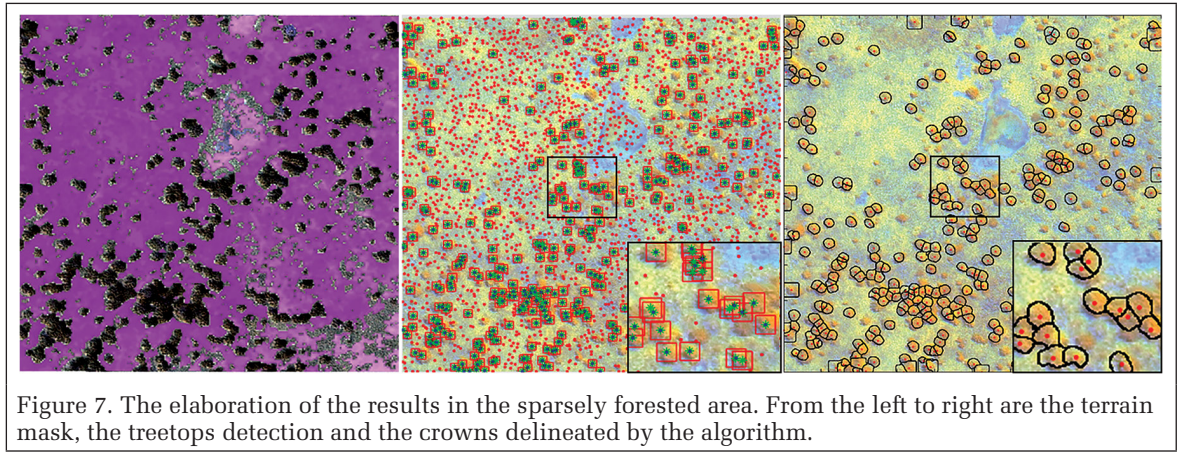

Figure 7. The elaboration of the results in the sparsely forested area. From the left to right are the terrain mask, the treetops detection and the crowns delineated by the algorithm.

As shown in Table 1 and Figure 7, at the sparsely forested area, the proposed algorithm predicted *355* trees comparing to *307* reference trees. The detection accuracy (*DA*) can reach as high as *0.89*. For the one-to-one correctly matched trees, their average overlap is *0.64*, which is better than the perfect match defined as *0.5* in Yin and Wang (2016).

Figure 8 shows examples of errors such as false negatives (diamond) and false positives (rectangle), as well as the miss-segmentations (circle), which are mainly caused by incorrectly detected treetops. The false negatives mainly exist in the short trees which are filtered by the above ground height check and non-maximum suppression that were previously described. Errors resulting in over-detection (circle in Figure 8) are mainly caused by the fact that grass near a short tree is confused as part of the crown.

In the densely forested area, the trees are highly overlapped with each other making it extremely difficult to distinguish individual crowns even by visual identification. As observed in Figure 9, the algorithm extracted the trees with a detection accuracy of *0.89* (Table 2). As compared to the sparsely forested area, the commission error ($e_{com}$) is relatively large, indicating relatively more false detections. This may due to that the dense canopies make the incorrectly detected treetops passed the postprocessing and identified as real trees. Furthermore, the low crown precision may be caused by partial detection and errors in the DSM. The partial detection refers to the detection that some branches of the tree are not included in the crown. For example, several treetops may be incorrectly detected in a large tree causing some branches could be misassigned to them. However, the post-processing may remove these incorrect detections and causes the true treetop only be assigned part of the crown. On the other hand, in the densely forested area, it is easy to mismatch feature points which are used to generate DSM from multi-view images. Hence, in this area, the 3D structure information for treetop detection may be not correct and further affect the superpixel segmentation.

Table 1. The experiment results in the sparsely forested area which has total *307* reference trees. Np refers to the number of the prediction. TH is the proposed top-hat treetop detector, while F_3, F_7, F_11, F_15, and F_19 represent the window filters with different sizes and SB stands for the variable window by slope breaks. The best numbers are bolded while other notations are the detection accuracy (*DA*), commission error ($e_{com}$), omission error ($e_{om}$) and crown accuracy (*CA*), precision (*P*) and F-scores (*F*).

| Detector | Np | DA | $e_{com}$ | $e_{om}$ | CA | P | F |
|---|---|---|---|---|---|---|---|
| TH | 355 | 0.89 | 0.23 | 0.12 | **0.64** | 0.76 | **0.82** |
| F_3 | 410 | **0.90** | 0.32 | 0.10 | 0.63 | 0.68 | 0.77 |
| F_7 | 386 | **0.90** | 0.28 | **0.09** | **0.64** | 0.72 | 0.80 |
| F_11 | 525 | **0.90** | 0.47 | 0.10 | 0.58 | 0.53 | 0.66 |
| F_15 | 355 | 0.83 | 0.27 | 0.16 | 0.62 | 0.72 | 0.78 |
| F_19 | 280 | 0.77 | 0.14 | 0.22 | **0.64** | 0.85 | 0.81 |
| SB | 219 | 0.67 | **0.05** | 0.32 | 0.56 | **0.95** | 0.79 |

Table 2. The experimental results in the densely forested area where has total *945* reference trees. The notations are the same as Table. 1.

| Detector | Np | DA | $e_{com}$ | $e_{om}$ | CA | P | F |
|---|---|---|---|---|---|---|---|
| TH | 1442 | **0.89** | 0.41 | **0.11** | **0.59** | 0.58 | 0.70 |
| F_3 | 349 | 0.24 | 0.34 | 0.76 | 0.58 | 0.66 | 0.36 |
| F_7 | 1193 | 0.82 | 0.35 | 0.18 | 0.58 | 0.65 | **0.73** |
| F_11 | 1330 | 0.83 | 0.41 | 0.17 | 0.58 | 0.59 | 0.69 |
| F_15 | 1009 | 0.74 | 0.30 | 0.22 | 0.58 | 0.70 | 0.72 |
| F_19 | 684 | 0.59 | 0.19 | 0.40 | 0.58 | 0.81 | 0.68 |
| SB | 543 | 0.51 | **0.12** | 0.50 | 0.59 | **0.88** | 0.64 |

Table 3. The experimental results in the urban area where has total *187* reference trees. The notations are the same as Table 1.

| Detector | Np | DA | $e_{com}$ | $e_{om}$ | CA | P | F |
|---|---|---|---|---|---|---|---|
| TH | 149 | **0.53** | 0.34 | 0.47 | 0.59 | 0.66 | **0.59** |
| F_3 | 194 | 0.52 | 0.49 | 0.48 | 0.58 | 0.51 | 0.51 |
| F_7 | 158 | 0.51 | 0.39 | 0.49 | 0.58 | 0.61 | 0.56 |
| F_11 | 170 | **0.53** | 0.41 | **0.46** | 0.58 | 0.59 | 0.56 |
| F_15 | 131 | 0.49 | 0.31 | 0.51 | 0.59 | 0.70 | 0.57 |
| F_19 | 107 | 0.42 | 0.27 | 0.58 | **0.61** | 0.73 | 0.53 |
| SB | 69 | 0.29 | **0.21** | 0.71 | **0.61** | **0.78** | 0.42 |

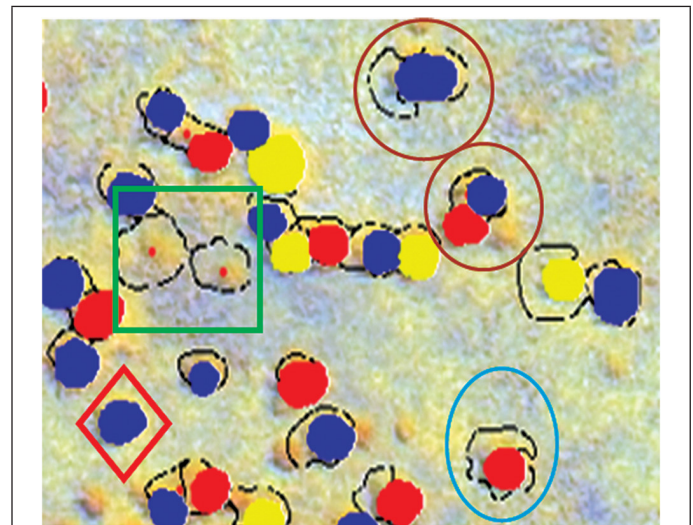

Figure 8. The presence of errors in the results.

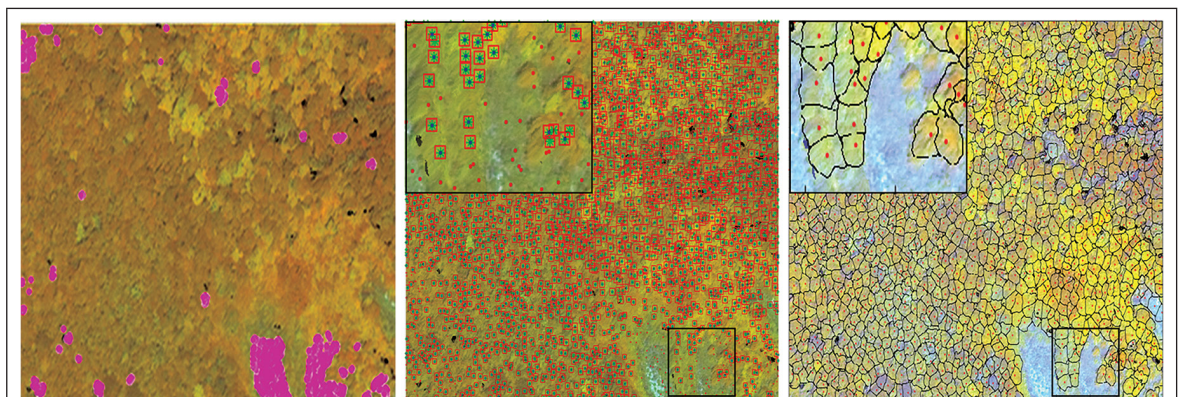

Figure 9. The elaboration of the results in the densely forested area. From the left to right are the terrain mask, the treetops detection, and the crowns delineated by the algorithm.

One consideration in this area is that the terrain area is too limited to offer an accurate estimation of the tree height. To investigate the situation without the DTM, we conducted two additional tests: (1) giving a constant height, here we use *10* meters (C_10), to all trees as an assumption., and (2) using the lowest point in the test area as the ground to calculate tree height. For the second test, we constrain the maximal tree height to avoid picking up the artifacts. Since the size of non-maximum suppression in the treetop refinement is related to the tree height and can further affect the final detections, we evaluated two maximal tree height constraints, *13* meters (L_13) and *15* meters (L_15).

The results of these tests can be found in Table 5, and they show that the performances in all cases are relatively consistent (the worst decrease of *DA* is *0.08* of L_15). For the constant height (C_10), the performance of detection accuracy (*DA*) and omission error ($e_{om}$) has even slightly increased (*0.01*) as this particular threshold results in more tree segments. However, some of these new trees are incorrect which increase the commission error ($e_{com}$) and decrease the precision. Comparing the next two tests that set the lowest point as ground, the L_13 has higher *DA* (*0.07*) than the L_15. From the Table 6 which presents the number of treetops after each processing, we can find the main difference between L_13 and L_15 is in the non-maximum suppression step which's filter size is dynamically determined by the tree height. From the results, we can infer that by subtracting the lowest point, plenty of trees have reached the maximal tree height and a higher maximal value offers a larger non-maximum suppression size thus filtered more potential trees. Comparing the DTM based height and the assumed height, the main difference is in the above ground height check previously described. Since the DTM can offer more accurate above ground tree height estimation, some lower trees or vegetation would be removed in the height check while with the assumed heights, all the treetops are passed. However, with a suitable tree height assumption, the abnormal low treetops can be eliminated by the non-maximum suppression such as the comparable performances shown in the Table 5. Hence, we can see that the proposed algorithm is able to maintain a high performance even without accurate DTM in the densely forested area.

At the test site C, the surface objects are complicated. However, as shown in Figure 10, the proposed algorithm still detected most of the trees even though they have different crown sizes and heights. With the help of 3D structural and spectral information, the proposed algorithm is able to distinguish the trees from confusing objects such as bushes and man-made objects. From Table 3, we can observe the $e_{com}$ is large (*0.47*) and from Figure 11 we can find that the trees with small crown size or nearby buildings are not detected. We believe this is mainly caused by the limited precision of the DSM. The trees that are too thin or too close to other objects may not be completely represented in the DSM, as Figure 11 shows below.

The resolution of the DSM is critical and the *0.3* meters GSD may be insufficient to represent the details of small trees. In addition, the generation may introduce geometric errors during the dense matching process, such as in specular reflection and textureless regions. Hence, we believe a better DSM can further improve the performance of the proposed method.

Table 4. The results with (S_3D) and without (S_2D) 3D information at the sparsely forested area (site A), densely forested area (site B), and the urban area (site C). The notations are the same as Table 1.

| Site | method | Np | DA | $e_{com}$ | $e_{om}$ | CA | P | F |
|---|---|---|---|---|---|---|---|---|
| A | S_3D | 355 | 0.89 | 0.23 | 0.12 | 0.64 | 0.76 | 0.82 |
|   | S_2D | 2439 | 0.70 | 0.91 | 0.30 | 0.49 | 0.09 | 0.16 |
| B | S_3D | 1442 | 0.89 | 0.41 | 0.11 | 0.59 | 0.58 | 0.70 |
|   | S_2D | 2053 | 0.84 | 0.61 | 0.16 | 0.52 | 0.39 | 0.53 |
| C | S_3D | 149 | 0.53 | 0.34 | 0.47 | 0.59 | 0.66 | 0.59 |
|   | S_2D | 143 | 0.18 | 0.76 | 0.82 | 0.53 | 0.24 | 0.21 |

Table 5. The detection results in the densely forested area with/without DTM information.

|  | DA | $e_{com}$ | $e_{om}$ | CA | P | F |
|---|---|---|---|---|---|---|
| DTM | 0.89 | 0.41 | 0.11 | 0.59 | 0.58 | 0.70 |
| C_10 | 0.90 | 0.52 | 0.10 | 0.58 | 0.48 | 0.63 |
| L_13 | 0.88 | 0.47 | 0.12 | 0.58 | 0.53 | 0.64 |
| L_15 | 0.81 | 0.39 | 0.19 | 0.58 | 0.61 | 0.69 |

Table 6. The numbers of treetops after each processing step.

|  | Initial | Height Check | Non-Maximum Suppression | Post-Processing |
|---|---|---|---|---|
| DTM | 2753 | 2491 | 1515 | 1442 |
| C_10 | 2753 | 2753 | 1801 | 1759 |
| L_13 | 2753 | 2753 | 1627 | 1588 |
| L_15 | 2753 | 2753 | 1278 | 1253 |

## Conclusions

In this paper, we developed a novel and automated method to fully utilize the multi-view high-resolution satellite images for ITDD. As compared to previous image-based methods, we adopt the DSM (digital surface model) derived from the multi-view satellite images and combine the multi-spectral information to identify treetops and their crowns in areas with varying canopy densities. A quantitative evaluation of three different sites shows that the proposed method is able to detect individual trees in different regions with various surface covers. The algorithm had its highest performance in the sparsely forested area with *89%* detection accuracy, *0.23* commission errors and *0.12* omission errors. Even for the densely forested area, traditionally deemed as particularly challenging, the algorithm still achieved *89%* detection accuracy with the slightly larger commission and omission errors.

Despite the superior results achieved by our methods, we are aware that significant vulnerabilities still exist, mainly due to the complicated surfaces of overly dense forests as well as propagated DEM errors. Detection in highly heterogeneous forests with multiple layers is challenging for even manual identification. The variations of trees and man-made objects in close proximity to the vegetation in urban areas create a very complicated scenario for individual tree detection. In the future, we plan to include dynamically adjusted tree templates of various scales in both 2D and 3D to increase the robustness to DEM errors and reduce overdetection in dense forest regions.

## Acknowledgments

The authors would like to thank John Hopkins University Applied Physics Lab for providing the Multi-view 3D Benchmark dataset used in this study. We would also thank Xing Pei and Ruopeng Wang from Lanzhou Jiaotong University for providing the tree labels. Finally, we thank jiaqiang Li from Future Cities Laboratory in Singapore-ETH Centre for the using of the CloudCompare® software for the terrain detection.

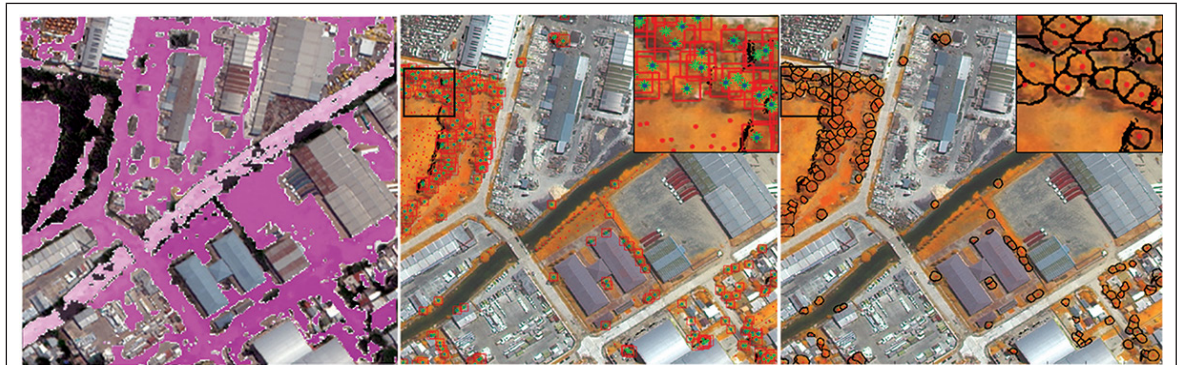

Figure 10. The elaboration of the results in the urban area. From the left to right are the terrain mask, the treetops detection and the crowns delineated by the algorithm.

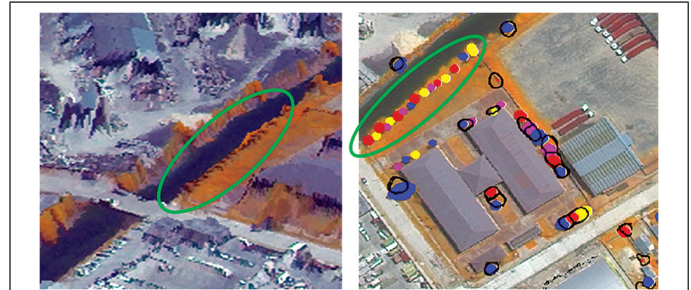

Figure 11. The mis-detection of thin trees. The left image is the 3D visualization of the test site and the right image is the detection results marked with reference trees. The ellipse marks the missing trees that only show as small bumps at the DSM.